\documentclass{article}
\usepackage{spconf,amsmath,amssymb,graphicx}
\usepackage{authblk}
\usepackage{tabularx}
\usepackage{booktabs}
\usepackage{microtype}
\usepackage{caption}
\usepackage{dblfloatfix}
\usepackage{bm}
\usepackage{hyperref}
\usepackage{enumitem}
\usepackage{setspace}
\usepackage{wrapfig}
\usepackage{filecontents}
\usepackage{etoolbox}
\newcolumntype{B}{<{\hspace{-1ex}}c}

\newcommand{\irow}[1]{
	\begin{matrix}(#1)\end{matrix}%
}
\newcommand{\ignore}[1]{}
\expandafter\def\expandafter\normalsize\expandafter{%
	\normalsize
	\setlength\abovedisplayskip{6pt}
	\setlength\belowdisplayskip{6pt}
	\setlength\abovedisplayshortskip{6pt}
	\setlength\belowdisplayshortskip{6pt}
}

\title{{\vspace{-0cm}Semi-Blind Spatially-Variant Deconvolution in Optical Microscopy\\with Local Point Spread Function Estimation By Use Of\\Convolutional Neural Networks}}

\name{Adrian Shajkofci$^{1,2}$, Michael Liebling$^{1,3}$\thanks{SNSF Grants 206021\_164022, 200020\_179217.}\thanks{\textcopyright 2018 IEEE. Personal use of this material is permitted. Permission from IEEE must be obtained for all other uses, including reprinting/republishing this material for advertising or promotional purposes, collecting new collected works for resale or redistribution to servers or lists, or reuse of any copyrighted component of this work in other works.}}
\address{$^{1}$Idiap Research Institute, CH-1920 Martigny, Switzerland\\
$^{2}$\'Ecole Polytechnique F\'ed\'erale de Lausanne, CH-1015 Lausanne, Switzerland\\
$^{3}$Electrical \& Computer Engineering, University of California, Santa Barbara, CA 93106, USA}

\begin{document}
%

\maketitle
\begin{abstract}
	
We present a semi-blind, spatially-variant deconvolution technique aimed at optical microscopy that combines a local estimation step of the point spread function (PSF) and deconvolution using a spatially variant, regularized Richardson-Lucy algorithm \cite{dey_richardsonlucy_2006}. To find the local PSF map in a computationally tractable way, we train a convolutional neural network to perform regression of an optical parametric model on synthetically blurred image patches.
We deconvolved both synthetic and experimentally-acquired data, and achieved an improvement of image SNR of 1.00 dB on average, compared to other deconvolution algorithms.

\end{abstract}
\section{Introduction}
\label{sec:intro}

Optical microscopy is a powerful tool to comprehend biological systems, enabling researchers and physicians to acquire qualitative and quantitative data about cellular function, organ development, or diseases. However, light traveling through any imaging system undergoes diffraction, which leads to image blur \cite{gibson_diffraction_1989}. This represents an intrinsic limit and the determining factor for the resolution of an optical instrument, and thus limits visual access to details. Indeed, the optical system only collects a fraction of the light emitted by any one point on the object, and cannot focus the light into a perfect image.\ignore{ Moreover, the imaging system can deviate significantly from a uniform response \textbf{!!HERE!!}, due to aberrations introduced by non-ideal lenses and misaligned components.} Instead, the light spreads into a three-dimensional diffraction pattern. Image formation can be modeled as the convolution of the original object with a PSF, which \ignore{is the image of an infinitely small point source in the object space and }sums up the optical aberrations \cite{sage_deconvolutionlab2_2017}. For thin, yet not flat, samples, the PSF remains shift-invariant within small areas of the 2D image, but the three-dimensional depth of the imaged object produces a local blur. Using a PSF corresponding to the blur in a deconvolution algorithm can be used to restore details in the image \cite{griffa_comparison_2010}.
\\
Deconvolution techniques can be categorized into three classes: (1) Non-blind methods, (2) entirely blind methods, and (3) parametric semi-blind algorithms. Non-blind methods require knowledge of the PSF \cite{soulez_learn_2014}. One of the main difficulties in practice is to determine the original PSF that characterizes the actual optical system without discovering it empirically by acquiring a 3D image of a fluorescent bead, which is a tedious and time-consuming calibration step. The two latter classes fall into blind deconvolution (BD) techniques, which improve the image without prior knowledge of the PSF, the object or other optical parameters. Entirely blind algorithms, such as \cite{hirsch_efficient_2010} are based the optimization of a penalty function or a maximum a posteriori (MAP) estimation of the latent image or kernel \cite{levin_understanding_2011}. However, these methods typically use strong constraints such as sharpness along the object edges and do not always generalize to unexpected or noisy types of data \cite{fish_blind_1995}, which are common in microscopy images. Also, many BD techniques are computationally expensive, especially for larger convolution kernels, and assume spatially invariant PSFs.\ignore{ Furthermore, the deconvolution output does not identify the image formation parameters and is usually constrained to a small kernel size.}\ignore{Moreover, due to their inherent dimensionality, optimization algorithms generally do not work in a real-time manner and cannot be used for live microscopy applications, such as live tracking or auto-focus.}
Finally, parametric or semi-blind algorithms are blind methods that are constrained by knowledge about the transfer function distribution, such as a diffraction model or a prior on the shape of the PSF (\cite{aguet_model-based_2008}, \cite{morin_semi-blind_2013}). Parametric models allow reducing the complexity of the optimization problem, increasing the overall robustness, and avoiding issues such as over-fitting. However, it remains hard to estimate the parameters from experimental data. We will focus on this third class of deconvolution methods, by addressing the central problem of how to best infer the parameters without measuring any of them experimentally.
\\
Machine learning recently improved the ability to classify images \cite{krizhevsky_imagenet_2012}, detect objects, or describe content \cite{girshick_rich_2014}. Convolutional Neural Networks (CNNs) \cite{lecun_gradient-based_1998}, in particular, are built for learning new optimal representations of image data and perform self-regulating feature extraction \cite{lecun_deep_2015}. Because of their ability to learn correlations between high- and low-resolution training samples, CNNs appear well adapted to our problem of determining the blur kernel. A similar reasoning led to recent results in \cite{sun_learning_2015} and \cite{nah_deep_2017}, where the direction and amplitude of motion blur was determined by a CNN classifier from images blurred with a Gaussian kernel.
\\
Here we present a spatially-variant BD technique aimed at microscopy of thin, yet non-flat objects. Our method combines local determination of the PSF and spatially-variant deconvolution using a regularized Richardson-Lucy (RL) algorithm \cite{dey_richardsonlucy_2006}. To find the PSF in a computationally tractable way, we train a CNN to perform regression of model parameters on synthetically blurred image patches. The novel aspects of our approach are: (1) Our method does not require the experimental measurement of a PSF, only synthetic training data is necessary. (2) Compared to non-parametric BD, the problem complexity remains low and therefore is more easily amenable to optimization. (3) Parameters with a physical meaning are inferred from the image itself. (4) The algorithm is computationally efficient, resulting in a near real-time kernel regression and mapping at the expense of a much longer, yet straightforward, training phase.
\\
In Section 2, we describe each step of the method. In Section 3, we present experiments with synthetic and acquired data. In Section 4, we conclude.

\section{Method}
\label{methods}





\begin{figure*}
	\begin{minipage}[b]{1.0\linewidth}
	  \centering
	  \centerline{\includegraphics[height=4.68cm]{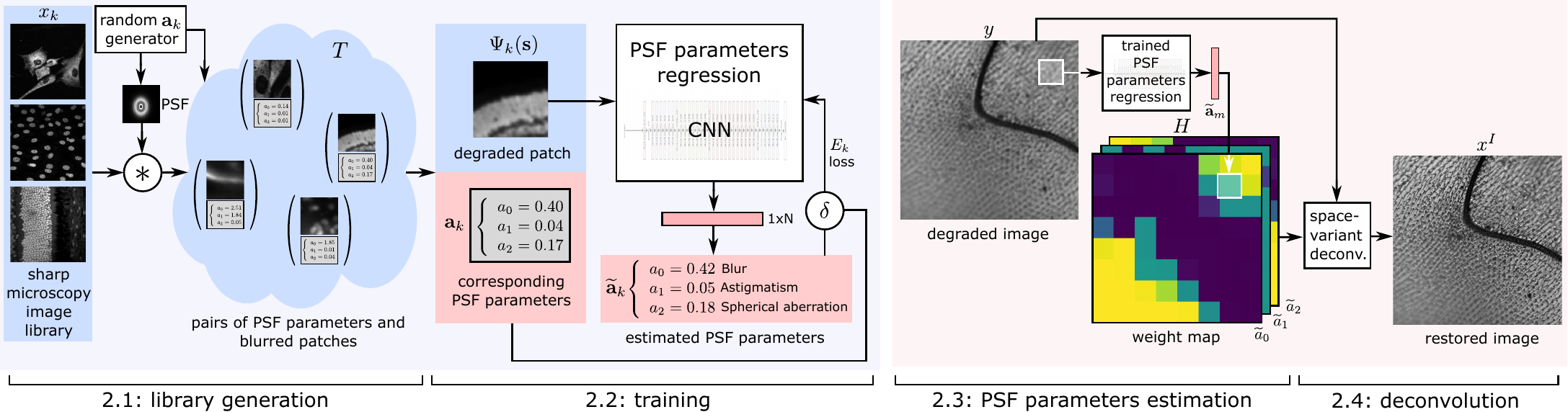}}
	  \caption{Left: method to estimate the parameters of the PSF that degraded an image patch. Right: method to estimate a map of the local PSF that degraded an image that can be fed into a spatially variant deconvolution algorithm.}
	  \label{fig1}
	\end{minipage}
\end{figure*}

We describe a deconvolution method for a 2D image in 3D space, such as a thin sample suspended in a gel. Blur in the acquired 2D image $y$ can be determined by an impulse response $h$.\ignore{The convolution of $h$a latent sharp image $x$} It is assumed that $h$ varies along the two lateral coordinates $\textbf{s}$ of the image. Thus, the optical system can be represented by the following image formation equation:
\begin{gather}
y(\mathbf{s}) = \iint x(\bm{\xi}) h(\mathbf{s},\bm{\xi}) \text{d}\bm{\xi}.
\end{gather}
In a nutshell, we propose the following steps to infer the parameters of the spatially-dependent function $h(\mathbf{s},\bm{\xi}) = h_{\mathbf{s}}(\bm{\xi})$:
\begin{enumerate}[align=parleft,leftmargin=11pt,labelsep=-8pt,topsep=6pt,itemsep=-1ex,partopsep=1ex,parsep=1ex]
	\item We develop a parametric model for $h_{\mathbf{s}}(\bm{\xi})$ allowing the generation of PSF/parameters pairs. We gather a training library of microscopy images, degraded by the space-invariant convolution of synthetic PSFs.
	\item We train a CNN on the above dataset to recognize the PSF model parameters given a unique blurred patch (Figure \ref{fig1}).
	\item Using a sliding window, we extract small patches from a large input image. For each patch, we regress the PSF parameters using the CNN. We combine the results and generate a map of PSFs over the input image.
	\item We recover a de-blurred image using a Total Variation (TV) regularized space-variant RL algorithm from the estimated map of PSFs.
\end{enumerate}
In the following subsections, we provide details on each of these four steps.
\subsection{Training library generation}
\label{sec:step1}
Optical abnormalities, such as defocus, astigmatism, or spherical aberrations, can be modeled with superpositions of Zernike polynomials in the expansion of the pupil function $W$ \cite{von_zernike_beugungstheorie_1934}:
\begin{gather} 
W_\mathbf{a}(\mathbf{s}) = \sum_{n=1}^{N} a_n  \cdot Z_n(\mathbf{s}), 
\end{gather}
where $\mathbf{s}$ denotes the two-vector of spatial coordinates, $N$ the maximal order of considered aberrations, and $\boldsymbol{a} = \irow{a_0&a_1&...&a_N}$ the parameters corresponding to each Zernike polynomial $Z_n$.
The pupil function can ultimately be converted, using a Fourier transform, into a PSF $h_{\mathbf{a}}(\mathbf{s}) = | \mathcal{F}(W_\mathbf{a}(\mathbf{s})) |^2$, with $\mathcal{F}$ the Fourier transform.
A training dataset is created by convolving $K$ images $x_k$, $k = 1, ... ,K$, with $K$ generated PSFs $h_{\mathbf{a}_k}$ of parameters $\mathbf{a}_k$ drawn from an uniform distribution. The degraded training images are obtained as follows:
\begin{gather}
\Psi_k(\mathbf{s}) = (h_{\mathbf{a}_k} \ast x_k)(\mathbf{s}) = \mathcal{F}^{-1}[\mathcal{F}(h_{\mathbf{a}_k})\mathcal{F}(x_k)](\mathbf{s}).
\end{gather}
We crop the images $\Psi_k$ to size $K_\psi \times L_\psi$ and pair them with their respective $\boldsymbol{a}_k$ in order to form the training set $T = \{(\psi_k, \boldsymbol{a}_k)\}_{k=1}^K$ (see Figure \ref{fig1}, part 2.1).
\\[3pt]
\textbf{Implementation:} We gathered a library of microscopy images from publicly available sources
, augmented with synthetic cell images. We blurred the images by $127 \times 127$ PSFs generated with parameters $a_\mathbf{k}$ drawn from an uniform distribution. We generated PSFs with $N=1$ and $N=2$. When $N=1$, the covered aberration was the focus, when $N=2$, the aberrations were the former plus the astigmatism.\ignore{The patches are of size $K_\psi = L_\psi = 128$.} The images were cropped to $128 \times 128$. We added Poisson noise and subtracted the pixel-wise mean from every patch. Images that did not comply with a minimal variance and white pixel ratio were rejected from the library. We applied image rotation to further augment $K$ to 2 million pairs.
\subsection{CNN training for PSF parameter regression}
\label{sec:step2}
Given an image patch as input, we wish to estimate the parameters of the PSF that was used to degrade the patch. For that purpose, we train a CNN for regression of these parameters from the established training set $T$ (Figure \ref{fig1}). We aim at training the neural network to learn the Zernike parameters that specify the the PSF by minimizing the Euclidean loss function:
\begin{gather}
E_k = \frac 1 {2N} \sum_{n=1}^N | {a_{k,n}} - {\widetilde{a}_{k,n}} |^2,
\end{gather}
where ${\widetilde{a}_{k,n}}$ is the current estimate of the $n$-th Zernike parameter on the training image $\psi_k$ and ${a_{k,n}}$ its ground-truth as stored in $T$.
\\[3pt]
\textbf{Implementation:} We trained two CNN models that we selected based on their good performance in the ImageNet competition: AlexNet \cite{krizhevsky_imagenet_2012}, adapted to gray-scale $128 \times 128$ input images, and a 34-layer ResNet \cite{he_deep_2016}. We modified the last fully-connected layer to have $N$ outputs. We trained both networks with the Adam optimizer in Caffe \cite{jia_caffe_2014} for 10 epochs. 

\ignore{\\
	\\We have now trained a model aimed at detecting the degradation parameters from a single image. In the following sections, we use the model for locally-variant PSF estimation and deconvolution.}
\subsection{Blind spatially variant kernel estimation of local PSF parameter map}
\label{sec:step3} 
Given the trained model able to recover the degradation parameters from a single image patch, we now turn to the problem of locally estimating the parameter of the PSF that degraded a larger input image. To achieve this, we use an overlapping sliding window over the input image with stride $t$ that is fed into the locally invariant regression CNN trained in \ref{sec:step2} (see Figure \ref{fig1}, part 2.3). We store the resulting parameters in a map $H = \{h_m\}_{m=1}^M$, where $h_m$ is the PSF corresponding to the patch $m$ with parameters $\mathbf{a}_m$ and $M$ the total number of patches and thus local PSF kernels.
\\[3pt]
\textbf{Implementation}: The window size was the same as the training patches $\psi_k$. \ignore{The final weighted and combined output spatial resolution was $ \lfloor \frac{K_x - K_\psi}{t}\rfloor+1 \times \lfloor \frac{L_x - L_\psi}{t}\rfloor+1$, with $K_x$, $L_x$, and $K_\psi$, $L_\psi$ being the width and height of the input image and the window, respectively.} For example, a $1024\times1024$ input image using $128\times128$ patches and $t=64$ yields $M = 169$ spatially-dependent PSF kernels. In order to avoid disparities and artifacts, we smoothened the parameter map with median filtering.

\subsection{Spatially variant deconvolution}
\label{sec:step4}
Given the degraded image and a local map of PSF parameters, we can now restore the input using TV-RL deconvolution. RL is an iterative maximum-likelihood approach derived from Bayes's theorem and assumes that the noise follows a Poisson distribution \cite{richardson_bayesian-based_1972}, which is well adapted for microscopy. The method is subject to noise amplification, which can, however, be counterbalanced by a regularization term that penalizes the $l_1$ norm of the gradient of the signal \cite{dey_richardsonlucy_2006}. Here, we assume that the PSF is spatially invariant in small parts of the image. We detail these steps in the sub-sections below.

\subsubsection{Spatially variant filtering}
\label{filtering}
\begin{figure}
	\begin{minipage}{1.0\linewidth}
		\centering
		\centerline{\includegraphics[width=8.6cm]{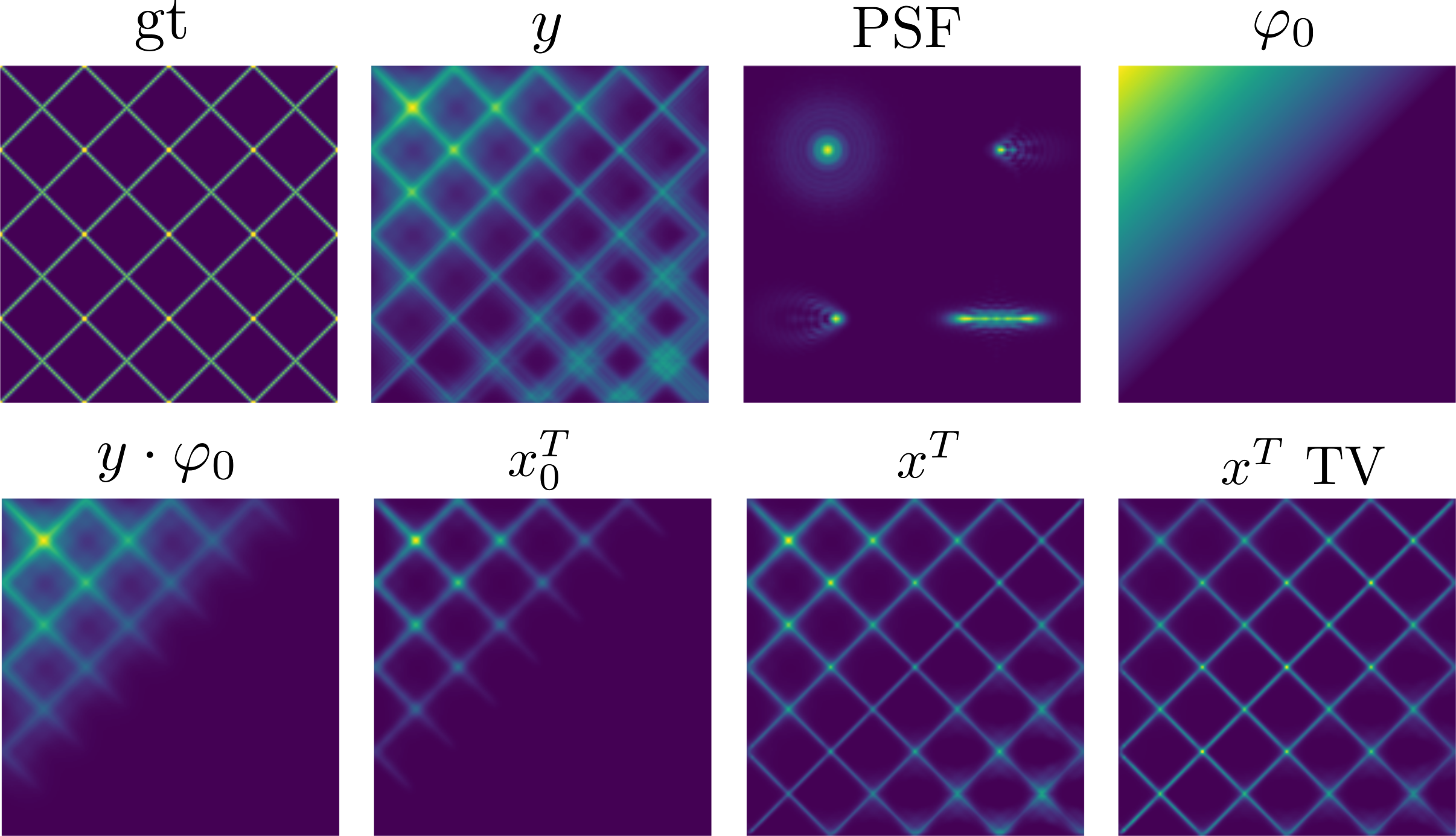}}
		\ignore{\caption{Illustration of variables shown in eq. \eqref{eq:eq7}, with $M=4$. Here, the ground truth is a synthetic grid as described in \ref{sec:synthdata}. $\varphi_m$ and $x_m^T$ are represented only when $m=0$.}}
		
		\caption{Synthetic experiment (\ref{sec:synthdata}). Starting from a ground truth image (gt), a local map of PSF (smooth interpolation between the 4 shown PSFs) and local weight combination $\bm{\varphi}$, we generated a blurred image y. The deconvolution method (similar to \cite{hirsch_efficient_2010}) starts from a map of locally estimated PSFs to deconvolve and recombine into a single image $x^T$ and $x^T$ TV.}
		\label{fig2}
	\end{minipage}
\end{figure}

Spatially variant convolution techniques have been extensively reviewed by Denis et al. (2015) \cite{denis_fast_2015}.\ignore{Nagy and O'Leary (1998) proposed to first convolve (FFT-based) the whole input with each PSF, and reconstruct the image using interpolation \cite{nagy_restoring_1998}.} Hirsch et al. have shown that our local invariance assumption can be improved by convolving the input with every local PSF and then reconstructing the image using interpolation \cite{hirsch_efficient_2010}.\ignore{Flicker and Rigaut followed a different perspective by decomposing the PSF into principal components \cite{flicker_anisoplanatic_2005}. Escande et al. used wavelet transforms to represent the optical transfer functions in a sparse basis \cite{escande_sparse_2015}; this method may improve windowing artifacts proper to spatially variant filtering.}
We extend this method\ignore{from \cite{hirsch_efficient_2010}} by using it as the basic convolution block in the TV-RL algorithm. Rather than interpolating deconvolved images, the method interpolates the PSF for each point in the image space \cite{temerinac_ott_spatially_2011}.
The idea for this pseudo space-variant filtering is: (i) to cover the image with overlapping patches using smooth interpolation, (ii) to apply to each patch a different PSF, (iii) to add the patches to obtain a single large image. The equivalent for convolution can be written as:
\begin{gather}
x(\mathbf{s}) = \mathcal{F}^{-1}\left[\sum_m^M \mathcal{F}(h_m(\mathbf{s})) \cdot \mathcal{F}(\varphi_m(\mathbf{s}) \cdot y(\mathbf{s}))\right],
\label{eq:eq5}
\end{gather}
where $\varphi_m(\mathbf{s})$ is the masking operator of the $m$th patch. We illustrated the masking and deconvolution steps in Figure \ref{fig2}.

\begin{table*}[b]
	\begin{minipage}{1.0\linewidth}
		\centering
		
		\small
		\begin{tabular}{cBBBBBBBBBBB} \toprule
			{$R^2_{\text{focus}}$} & {$R^2_{\text{focus, ast}}$} & {$\text{SNR}_\text{degraded}$} & {$\text{SNR}_\text{ours}$} & {$\text{SNR}_\text{\cite{holmes_light_1995}}$} & {$\text{SNR}_\text{\cite{kotera_blind_2013}}$} & {$\text{SNR}_\text{\cite{whyte_deblurring_2014}}$} & {$\text{SSIM}_\text{degraded}$} & {$\text{SSIM}_\text{ours}$} & {$\text{SSIM}_\text{\cite{holmes_light_1995}}$} & {$\text{SSIM}_\text{\cite{kotera_blind_2013}}$} & {$\text{SSIM}_\text{\cite{whyte_deblurring_2014}}$} \\ \midrule
			
			0.91 & 0.58 & {1.90  dB} & {\textbf{4.48 dB}} & {3.48 dB} & {1.4 dB} & {1.54 dB} & {0.10} & {\textbf{0.51}} & {0.28} & {0.29} & {0.52}\\ \bottomrule
		\end{tabular}
		\caption{Image recovery benchmark with 100 synthetic input images. We computed $R^2$, between the generated PSF parameters and the recovered PSF parameters, as well as SNR and SSIM.\ignore{The methods reviewed are Holmes et al., 1995 \cite{holmes_light_1995}, Kotera et al., 2013 \cite{kotera_blind_2013}, and Whyte et al., 2014 \cite{whyte_deblurring_2014}.}}
		\label{table:table1}
	\end{minipage}
\end{table*}
\subsubsection{Regularization of the RL algorithm}
In noisy images, the RL algorithm tends to exacerbate small variations. TV regularization is used to obtain a smooth solution while preserving the borders \cite{dey_richardsonlucy_2006}. The resulting image at each RL iteration becomes:
\begin{multline}
x^{i+1}(\mathbf{s}) = \sum_m^M
\left[\frac{(h_m * (\varphi_m \cdot y))(\mathord{\cdot})}{(h_m * x_m^t)(\mathord{\cdot})}
* {h_m}(-\mathord{\cdot})\right](\mathbf{s})
\\
\cdot
\frac{x_m^i(\mathbf{s})}
{1 - \lambda_{TV} \hspace{2pt}\text{div}\left(\frac{ \triangledown x_m^i(\mathbf{s})}{\lvert\triangledown x_m^i(\mathbf{s}) \rvert}\right)},
\label{eq:eq7}
\end{multline}
with $y$ the blurry image, $x^i$ the deconvolved image at iteration $i=1,...,I$, $x_m^i$ the $K_y \times L_y$ deconvolved patch at iteration $i$, $M$ the number of patches (and different PSFs) in one image $x$, $h_m$ the $K_h \times L_h$ PSF for patch $m$ and $\lambda_{TV}$ the regularization factor. $\triangledown x_i^j(s)$ is the finite difference operator, which approximates the spatial gradient.
\\[3pt]
\textbf{Implementation:} Here, $\varphi_m$ was a bilinear interpolating function.\ignore{ For optimization purposes, we resized the masked image to its non-zero part.} If $\lambda_{TV}$ is small, RL is dominated by the data, and if $\lambda_{TV} \sim 1$, RL is dominated by the regularization \cite{dey_richardsonlucy_2006}. Therefore, we used a balanced $\lambda_{TV} = 0.001$.
Since using FFT-based calculations implies that $h$ is circulant, we had to take into account the field extension to prevent spatial aliasing.\ignore{ The data was padded with zeros and was extended to $(K_h+K_y-1) \times (L_h+L_y-1)$.} The number of RL iterations, $I$, was fixed to $20$.

\section{Experimental results}
\label{sec:results}
We validated the method using synthetic and live microscopy experiments.\ignore{ After having trained the model as described in \ref{sec:step2}, we obtained a spatially variant PSF parameters map (\ref{sec:step3}). Finally, we deconvolved the input image using TV-RL (\ref{sec:step4}).} The results were compared using the SNR and SSIM \cite{wang_image_2004} with three BD methods : 1. MATLAB's \texttt{deconvblind} \cite{holmes_light_1995}; 2. the non-CNN deconvolution method \cite{kotera_blind_2013}, and 3. the space-variant BD described in \cite{whyte_deblurring_2014}. Due to gain difference between methods and boundary effects, we computed the maximum SNR obtainable following linear scaling and offset of the gray values.

\subsection{Synthetic data}
\label{sec:synthdata}
We generated a $252 \times 252$ gray-scale grid pattern, as shown in Figure \ref{fig2}. Using the technique in equation (\ref{eq:eq5}), we degraded each quadrant of the input image with a specific, randomly-generated $127 \times 127$ PSFs using parameters $\mathbf{a}$ drawn from a uniform random distribution. We subsequently inferred a PSF map via the CNN from the blurry image as described in \ref{sec:step3}. Finally, we deconvolved the image and determined the match $R^2$ between the Zernike parameters used for PSF generation and the recovered parameters, as well as the reconstruction quality. Results in Table \ref{table:table1} indicate an  improvement of the spatially variant BD in comparison to invariant BD.

\subsection{Acquired data}
\begin{figure}
	\begin{minipage}{1.0\linewidth}
		\centering
		\begin{tabular}{cB}
			{Original image} & {Deconvolution \cite{whyte_deblurring_2014}} \\
			{\includegraphics[width=2.6cm]{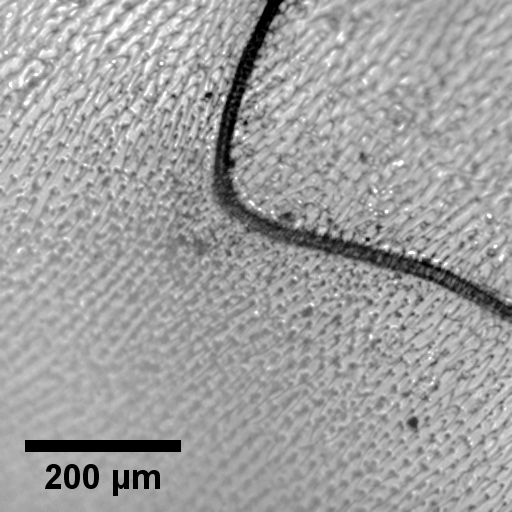}} & {\includegraphics[width=2.6cm]{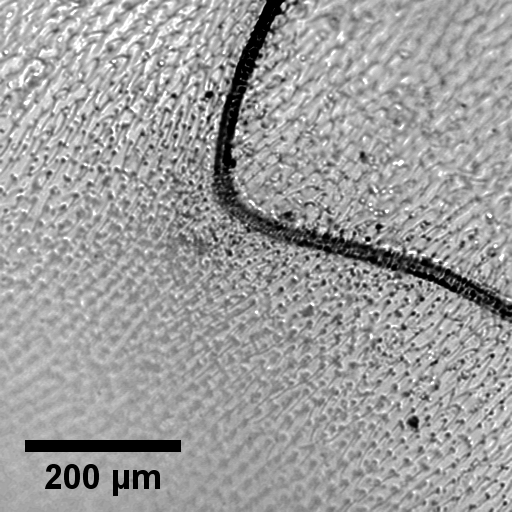}} \\
			{Focus ($a_0$) map} & {Deconvolution (our method)} \\
			{\includegraphics[width=2.6cm]{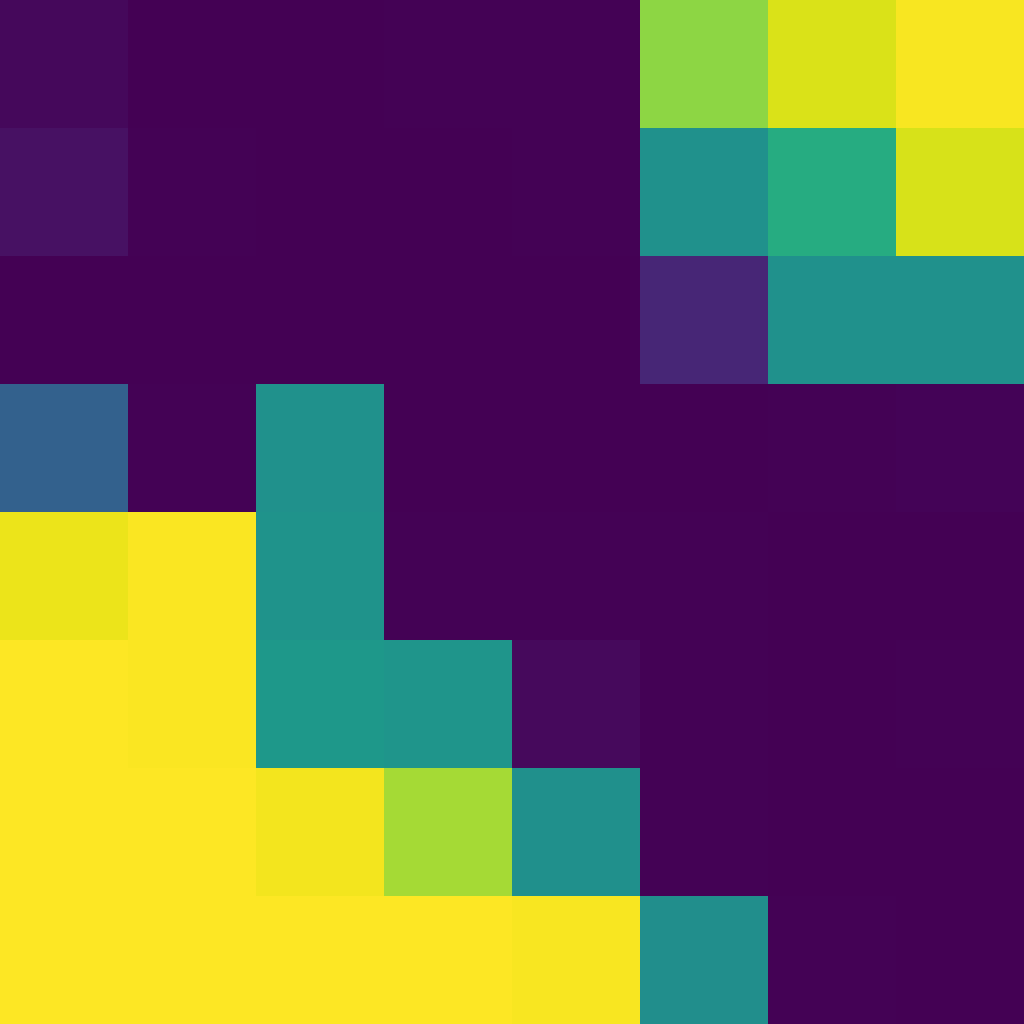}} & {\includegraphics[width=2.6cm]{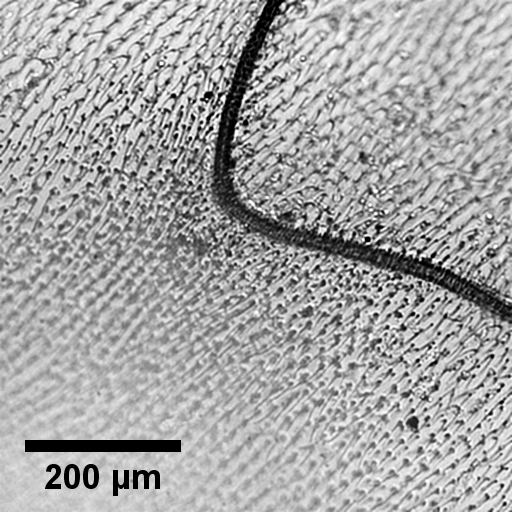}}
		\end{tabular}
		\caption{Deconvolution of a tilted housefly wing using our algorithm and the method from \cite{whyte_deblurring_2014}. Our pipeline inferred a $8 \times 8$ map of $\mathbf{a}$ coefficients. Here, a yellow color represents $a_0 = 2$, while a dark blue color represents $a_0 = 0$.}
		\label{fig3}
	\end{minipage}
\end{figure}

We used a wide-field microscope to acquire the image of a housefly wing. The sample was tilted 20 degrees so that only the central part of the image was in focus. The goal was to recover the in- and out-of-focus PSFs in order to restore the whole image.\ignore{ Then, we converted the defocus Zernike parameter into a depth map and calibrated it to a physical distance .} Our method gives visual results that are similar to \cite{kotera_blind_2013} (Figure \ref{fig3}). With $M=64$ regions, we generated the map $H$ in 18 milliseconds with a consumer-grade GPU. The additional time necessary for determining the PSF maps is negligible compared to the time necessary for deconvolution. 


\section{Discussion}
\label{sec:discussion}

Space-varying BD is a more complex process than space-invariant deconvolution. However, our experiments showed that it is possible to add parametric constraints to the problem and thereby allowing to map patches of spatially-invariant PSFs onto the input image.
With our CNN-based method, we were able to detect their original blur kernel with an average accuracy of 91.1\% (Table \ref{table:table1}), given only synthetic images as the training input (no experimental measurement of the PSF was necessary). From these results, we have been able to deconvolve these image with an SNR on average 1.00 dB higher than other blind deconvolution techniques. We used the method on acquired experimental data and observed a visual improvement similar or, in some regions, better than when we used other methods. The main limitation of our approach is that the number of regressed Zernike polynomials has an important influence on the performance. For example, the parameter ``focus'' is well recovered, but other parameters, such as coma, have a very low regression accuracy. We will focus future efforts into researching a CNN architectures that are more specific to the problem. In contrast to end-to-end deblurring methods such as \cite{wieschollek_learning_2017}, users may be able to easily customize our algorithm with other parametric models, CNN networks or deconvolution techniques for specific applications. Our method is therefore suitable to other applications than image enhancement, such as auto-focus or depth determination from focus \cite{bove_entropy-based_1993}. But mostly, our method could be used to simplify procedures that required physical measurements of PSFs such as \cite{quirin_optimal_2012}, where deconvolution improved PALM/STORM microscopy using PSFs from a set of calibration images.
\\
\\
Code and documentation are available at this address: 
\\
\href{https://github.com/ashajkofci/semiblindpsfdeconv}{https://github.com/idiap/semiblindpsfdeconv}.

\section{References}
\bibliographystyle{IEEEbib}
\small{
\begin{spacing}{0.9}
	
\bibliography{IEEEabrv,refs}
\end{spacing}
}
\end{document}